\documentclass[runningheads]{llncs}

\usepackage{eccv}

\usepackage{eccvabbrv}

\usepackage{graphicx}
\usepackage{booktabs}

\usepackage[accsupp]{axessibility}  % Improves PDF readability for those with disabilities.

\usepackage{hyperref}

\usepackage{orcidlink}

\usepackage{multirow}
\usepackage{graphicx}      % 이미지 삽입
\usepackage{caption}       % 캡션 제어
\usepackage{amsmath,amssymb}
\usepackage{booktabs,makecell}
\usepackage{stfloats}
\usepackage{array}
\usepackage{pgfplots}
\pgfplotsset{compat=1.18}
\usepackage[ruled]{algorithm2e}
\usepackage{subcaption}
\usepackage{placeins}

\begin{document}

\title{Q-TriM: Question-Guided Tri-Modal Attention for
Audio–Visual Question Answering} 
\titlerunning{Q-TriM}



\author{
\begin{tabular}{c@{\hspace{1.2cm}}c}
\begin{tabular}{c}
SungHun Kim\\
Dept. of Computer Science\\
and Engineering\\
Korea University\\
Seoul, South Korea\\
\texttt{sainthun12@korea.ac.kr}
\end{tabular}
&
\begin{tabular}{c}
SeungJun Baek\thanks{Corresponding author.}\\
Dept. of Computer Science\\
and Engineering\\
Korea University\\
Seoul, South Korea\\
\texttt{sjbaek@korea.ac.kr}
\end{tabular}
\end{tabular}
}

\authorrunning{S.~Kim and S.~Baek}

\institute{}

\maketitle

\begin{abstract}
Audio-Visual Question Answering (AVQA) extends classical VQA by requiring joint reasoning over video and synchronized audio. However, many AVQA systems rely on deeply stacked layers of self- and cross attention across text, video, and audio. 
%Such depth can amplify stage-wise information loss, since cues dropped early cannot be recovered and fusion errors accumulate across sequential attention blocks.
Such sequential stacking may incur loss of information such as subtle inter-modal cues over the layers, causing errors to accumulate across sequential attention layers during the fusion.
\color{black}
We introduce Q-TriM which performs multi-modal fusion in a shallow and parallel manner instead of a deep and sequential manner. For Q-TriM, we propose a novel framework for attention operation incorporating video and audio conditioned on text. As a result, we obtain not only standard cross attention outputs but also Tri-Modal Attention representations in which Query, Key, and Value come from distinct modalities.  
These attention representations are combined in parallel at a single stage, thus avoiding the multi-modal fusion with deep stacks in order to mitigate error accumulation and depth-induced issues. 
Q-TriM achieves state-of-the-art performance on three AVQA benchmarks, including substantial gains on MUSIC-AVQA-R, which demonstrates its robustness and out-of-distribution generalization. Code is available at \color{RubineRed}https://github.com/Sunghun95/Q-TriM\color{black}

\end{abstract}

\section{Introduction}
\label{sec:intro}

\begin{figure}[t]
  \centering
  \includegraphics[width=0.6\linewidth]{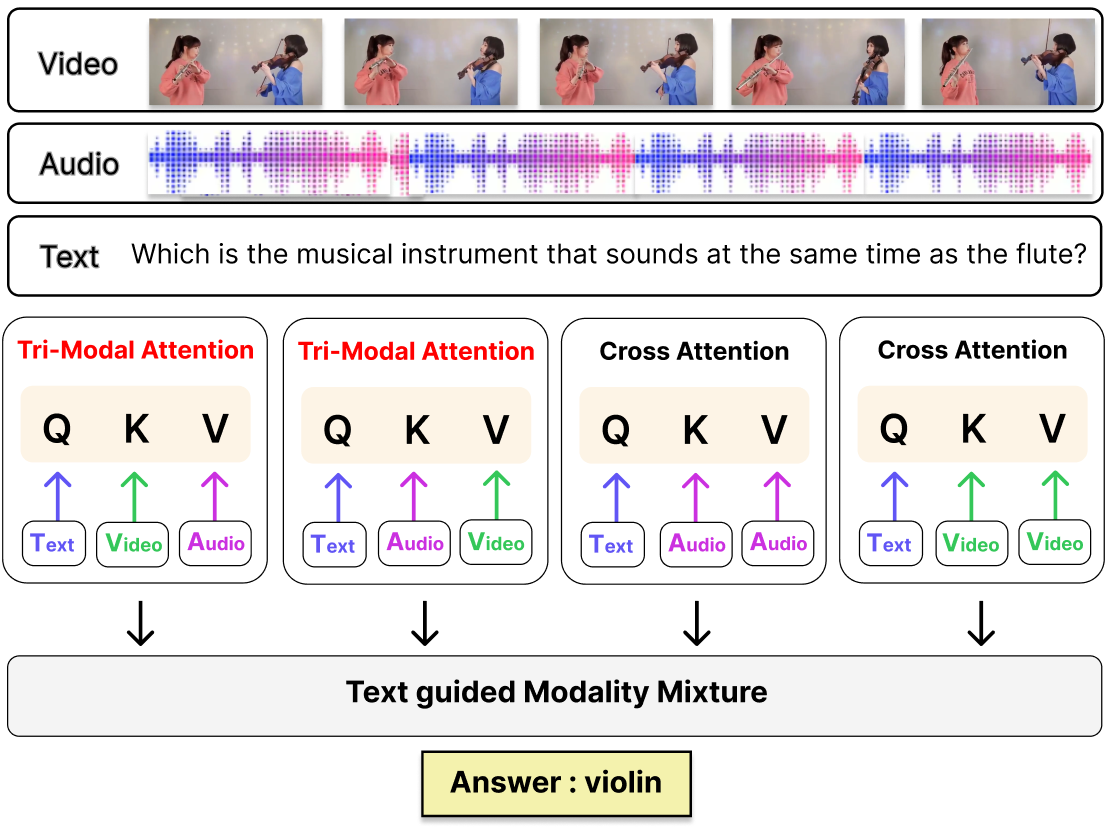}

  % 캡션: 라벨 옆에서 시작 + 다음 줄도 전체 폭으로 이어짐
  \captionsetup{
    format=plain,           % 행걸이(hang) 끔
    labelsep=period,        % "Figure 1." 형태
    justification=justified,
    singlelinecheck=false,
    width=\linewidth        % (템플릿에 따라 필요) 캡션 폭을 칼럼 폭으로
  }

  \caption{\textbf{Overview of Q-TriM.} Given input video, audio, and text, Q-TriM combines representations from Tri-Modal Attention, together with cross attention, to perform a shallow, parallel fusion of the modalities for reasoning of the answer.}
  \label{fig:overview}
\end{figure}

AVQA(Audio-Visual Question Answering) is the task of answering a natural-language question given a video with synchronized audio, extending classical VQA \cite{VQA} by jointly leveraging acoustic and visual streams \cite{AVQA}. AVQA spans many use cases: (1) scene-aware dialog/assistive agents that reason over live audio–video streams \cite{AVSD}, (2) long-form egocentric “memory” systems that find evidence moments for open-ended questions \cite{EMQA}, and (3) first-person action settings where EPIC-Fusion aligns sounds with egocentric visuals before answering \cite{EPICF}.

In AVQA, extracting salient evidence from both video and audio is essential. Most AVQA systems rely on \emph{stacked} attention layers across text, video, and audio \cite{QATIGER,MCAN,LAVISH,TSPM} and deepen the attention stack to capture inter-modal dependencies. In general, neural layers perform the abstraction and compression of information. On the one hand, these operations are necessary to selectively extract essential information, e.g., Information Bottleneck (IB) principle \cite{IB}. On the other hand, they accompany the loss of information \cite{IL1,IL2,IL3}, which may be problematic for fusing information from many modalities. For example, a modality processed in earlier layers may lose important inter-modal cues, making them unavailable for fusion at deeper layers. A simple remedy is to design architectures that achieve similar effects with shallow attention layers. However, since AVQA must integrate three modalities, stacking attention is often unavoidable. In standard cross-attention \cite{attention}, only two modalities interact at a time; a Query attends to a single Key–Value source, and thus at most two modalities interact simultaneously.  For example, prior methods \cite{TSPM,PSTP-NET} first process text and video with cross-attention, and then combine its output with audio representation via another cross-attention. Thus, coordinating text, video, and audio typically requires multiple fusion stages in a sequential manner.

In this paper, we propose \textbf{Q-TriM} (Question-guided Tri-Modal Attention and Modality Mixture) as illustrated in Figure~\ref{fig:overview}. To avoid stacked attention, we go \emph{shallow} and \emph{parallel} rather than \emph{deep} and \emph{sequential}. Instead of sequential layering of blocks, we adopt a parallel fusion step to compute mixed features across all three modalities. To this end, we propose a novel framework for multi-modal attention. Our framework induces a joint distribution of video and audio representations per frame conditional on the input text. The fused representation is computed as a conditional expectation from the distribution.
%We first form a joint embedding, computes a question-conditioned similarity score, and then derives expectation-based embeddings as outputs. 
Unrolling these expectations (see Sec. \ref{sec:attn} for details) reveals not only standard cross-modal attention but also a new attention form in which the query, keys, and values are drawn from different modalities; we refer to this operator as \textbf{Tri-Modal attention}. 
Importantly, Tri-Modal attention models the three-way interactions among text, video, and audio in a \emph{direct} manner within a single stage, thereby avoiding potential loss of inter-modal cues during the fusion.
Finally, we compute a question-aware mixture of these outputs to produce the most relevant fused representation for reasoning. In addition, to ensure that only question-relevant patches participate in learning, we propose a \emph{question-guided token filtering} module. Without introducing any extra attention, the token filtering module uses simple dot-product scoring and a straight-through estimator (STE) \cite{STE}, thereby preventing further deepening of the attention stack. Consequently, our method achieves state-of-the-art performance on the standard AVQA benchmarks, MUSIC-AVQA\cite{MUSIC-AVQA} and MUSIC-AVQA-R\cite{MUSIC-AVQA-R}. On MUSIC-AVQA-2.0\cite{MUSIC-AVQA-2.0}, it further attains SoTA in three out of four evaluation categories.

Our main contributions are summarized as follows:
\begin{itemize}
    \item We present \textbf{Q-TriM}, an AVQA model that performs inter-modal attention in a shallow, parallel fusion stage rather than by deep, sequential stacks. By consolidating fusion horizontally, Q-TriM achieves the effect of deeper cross-modal pipelines while curbing error propagation through deep layers. % and 
    % alleviating depth-induced issues such as over-smoothing and modality collapse.
    % \color{red}
    % while supporting question-guided evidence selection and making tri-modal interactions explicit within a single fusion stage.
    % \color{black}
    
    \item We propose a new modality-fusion framework that produces a question-conditioned joint distribution of video and audio frame tokens. Within this framework, we derive a Tri-Modal attention operator that allows distinct Query, Key, and Value sources across modalities, enabling direct and explicit three-way fusion without resorting to deep or sequential stacks.
    
    \item Q-TriM achieves state-of-the-art performance across three representative AVQA benchmarks. Notably, on the MUSIC-AVQA-R dataset \cite{MUSIC-AVQA-R}, Q-TriM achieves superior performance over baselines, demonstrating its robustness.
\end{itemize}

\color{black}

\section{Related Work}
We group prior Audio–Visual Question Answering (AVQA) research into three axes: question awareness, spatial perception, and temporal perception. These axes respectively target how text guides reasoning, where to look and listen in the scene, and when salient evidence occurs across time.% Most models combine at least one of these.

\subsection{Question Awareness}
In AVQA, it is important to condition the model on the question so that both feature extraction and fusion remain text-guided. Co-attention and question-guided fusion from VQA (e.g., \cite{MCAN,HCAttn,PSAC}) inspire AVQA systems that ground sounding regions and fuse audio–visual cues under textual guidance (e.g., \cite{MUSIC-AVQA,COCA,APL}). Recent work further strengthens conditioning via prompt reformulation or question-aware experts: TSPM turns questions into declaratives to align with visual semantics before temporal/spatial selection \cite{TSPM}, and \cite{QATIGER} introduces question-aware Gaussian experts. Dialog-style conditioning also appears in scene-aware settings \cite{AVSD}. Overall, these methods filter modalities through question semantics to suppress irrelevant content and to guide cross-modal reasoning \cite{TSPM,MUSIC-AVQA,LAVISH}.

\subsection{Spatial Perception}
Another line of work localizes question-relevant regions and sound sources, often by coupling detection with cross-modal grounding. Panoramic grounding requires spherical spatial reasoning in 360\textdegree~videos \cite{PANO-AVQA}. Progressive pipelines (e.g., PSTP-Net) first select key segments and then pick relevant patches with audio-guided visual attention \cite{PSTP-NET}. Complementary strategies include object-aware learning \cite{APL}, causal regularization against spurious cues \cite{COCA}, and parameter-efficient ViT-based learners \cite{LAVISH}. In practice, many approaches pair these spatial modules with question-aware fusion to maintain textual guidance throughout \cite{MUSIC-AVQA,TSPM}.

\subsection{Temporal Perception}
Temporal perception targets frames or segments that matter for the question, ensuring that evidence is aggregated over the correct intervals. PSTP-Net explicitly performs segment selection before spatial filtering \cite{PSTP-NET}, while TSPM adds a prompt-guided temporal module \cite{TSPM}. Earlier video-/audio-QA studies explore temporal reasoning with recurrent or attention-based models (e.g., \cite{HCRN,PSAC,FCNLSTM}), and dialog-style AV understanding leverages temporal context as well \cite{AVSD}. AVQA systems often integrate temporal selection with spatial grounding and question-aware fusion for end-to-end reasoning \cite{MUSIC-AVQA}.

Overall, inter-modal attention is typically stacked in the aforementioned works (at least two layers). Notably, QA-TIGER \cite{QATIGER}, the state-of-the-art method at the time of writing, uses four cross-attention blocks which are sequentially connected. %, raising depth-related risks such as modality collapse and over-smoothing in fusion heads.

\section{Method}

\begin{figure*}[t]
  \centering
  \includegraphics[width=\textwidth]{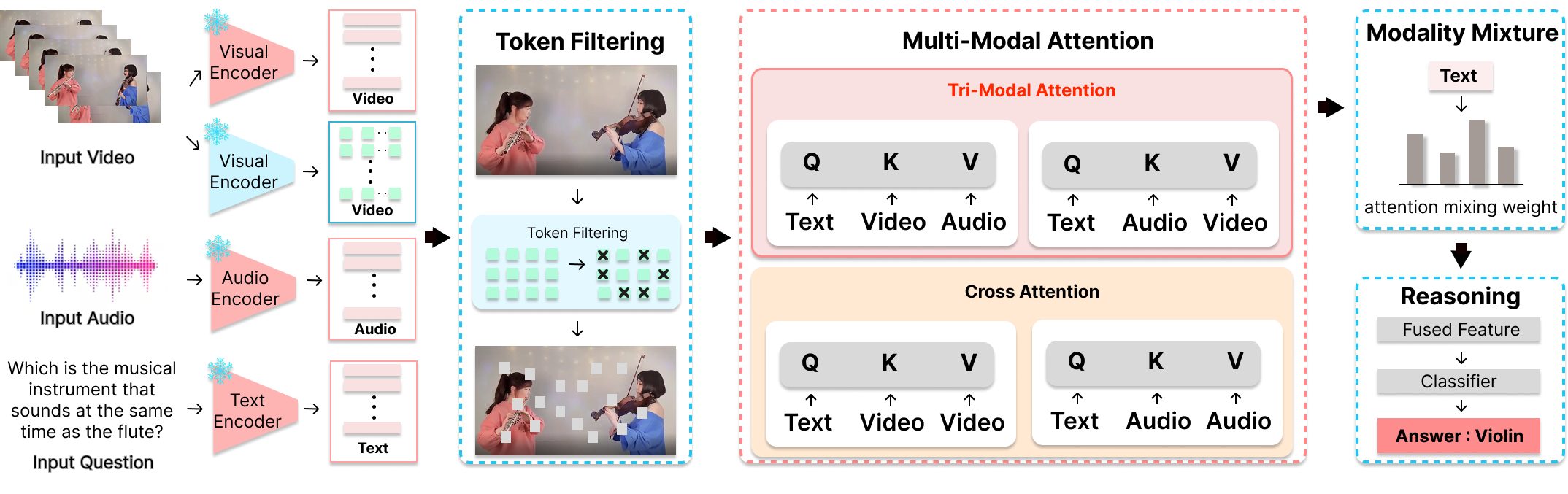}

  % 캡션: "Figure 2." 라벨과 문장이 같은 줄에서 시작, 다음 줄도 전체 폭으로 이어짐
  \captionsetup{
    format=plain,
    labelsep=period,
    justification=justified,
    singlelinecheck=false,
    width=\textwidth
  }

  \caption{\textbf{Overall architecture of \textsc{Q-TriM}.} First, question-guided token filtering keeps only the top tokens most relevant to the text and drops the rest. Next, Question-aware modality mixture computes four paths: TMA(Text,Audio,Video), TMA(Text,Video,Audio), CA(Text,Video,Video), and CA(Text,Audio,Audio). TMA stands for Tri-modal attention, and CA stands for cross-attention. Then question-conditioned weights are applied to the outputs. Finally, fused-feature reasoning combines these outputs and feeds the result to a classification layer to produce the answer.}
  \label{figure2}
\end{figure*}

Q-TriM adopts a shallow and parallel architecture that combines cross-attention with Tri-Modal attention, instead of stacking attention blocks deeply. An overview of the architecture is shown in Figure~\ref{figure2}.

~\\\noindent\textbf{Notations.} A vector $\mathbf{v}$ is defined as a \emph{row} vector by default. The dot-product between vectors $\mathbf{u}$ and $\mathbf{v}$ is thus $\mathbf{u}\mathbf{v}^\top$. A matrix $\mathbf{X}=\{\mathbf{x}_i\}_{i=1}^N\in\mathbb{R}^{N\times D}$ can be viewed as having $N$ rows of $D$-dimensional vectors $\mathbf{x}_i$, $i=1,\dots,N$. Define the conventional dot-product attention for representations $\mathbf{Q},\mathbf{K},\mathbf{V}$ of proper dimensions\footnote{The attention originally contains normalization in the softmax, e.g., $QK^\top$ is scaled by $1/\sqrt{D}$ if the dot-product is of dimension $D$. We omit $1/\sqrt{D}$ for notational simplicity, and assume that such normalization is performed whenever necessary for implementation.}:
\begin{align}
\text{Attn}(\mathbf{Q}, \mathbf{K}, \mathbf{V}) := \text{softmax}\left(\mathbf{Q}\mathbf{K}^\top\right)\mathbf{V}\label{eq:attn}
\end{align}

\subsection{Input Representation}
%Because our design uses few attention layers, starting from disjoint embeddings would restrict cross-modal sharing. To mitigate this, we initialize text, video, and audio in a \emph{common} embedding space using ImageBind~\cite{imagebind}, so training begins from an aligned representation.
% We begin by projecting text, video, and audio in a \emph{common} embedding space using ImageBind~\cite{imagebind}. Such aligned representation can help cross-modal sharing of information and reduce their sequential processing. %This helps our training begins from an aligned representation.
We project three modalities, i.e., text, video, and audio, into a \emph{common} embedding space. Such aligned representations help modeling direct interactions across modalities via attention. Firstly, we project text and video into CLIP embedding spaces \cite{CLIP}. Secondly, we align audio with text and video; there are candidate methods extending CLIP to audio, e.g., AudioCLIP \cite{AudioCLIP}, ImageBind \cite{imagebind}. We choose a recent method: ImageBind. %, aligning it with the same space to enable subsequent attention-based interactions across modalities.
% We begin by projecting text and video into a \emph{common} embedding space using CLIP embeddings. We embed audio using ImageBind~\cite{imagebind}, aligning it with the same space to enable subsequent attention-based interactions across modalities.
%Such aligned representation can help cross-modal sharing of information and reduce their sequential processing.
\color{black}
We segment each original video into \(T\) contiguous \emph{2-second} clips and extract frame-aligned video and audio representations per segment. The text consists of a single question for the entire video. The details of input representation is explained as follows.

~\\
\noindent\textbf{Text representation.}\label{sec3.1}
%Unlike the frame-level signals above, 
The question text $\mathbf{t}$ is encoded as a single sentence-level embedding from CLIP \cite{CLIP} given by \[
\mathbf{t} \in \mathbb{R}^{D}.
\] %Note that $\mathbf{t}$ has no temporal dimension.

\noindent\textbf{Video representation.} We extract frame-level global video embeddings from CLIP given by \[
\mathbf{V}
= \{\,\mathbf{v}_{t}\,\}_{t=1}^{T}
\in \mathbb{R}^{T \times D}.
\]
Additionally, in order to retain spatial information, we obtain patch-level tokens using CLIP model given by \[
\mathbf{V}^{\mathrm{p}} = \{\,\mathbf{v}^{\mathrm{p}}_{t}\,\}_{t=1}^{T}
\in \mathbb{R}^{T \times P \times D}.
\]

\noindent\textbf{Audio representation.} %\color{red} While text and video are embedded with CLIP, \color{black}
We extract frame-level audio embeddings aligned with text and image CLIP using ImageBind \cite{imagebind} given by \[
\mathbf{A} = \{\,\mathbf{a}_{t}\,\}_{t=1}^{T}
\in \mathbb{R}^{T \times D}.
\]

\subsection{Token Filtering}
\noindent\textbf{Overview.} The entire visual representation $\mathbf{V}^{\mathrm{p}}$ from Sec.~\ref{sec3.1} may contain irrelevant or redundant tokens \cite{patch,patch2}. This can incur unnecessary computational overhead and produce noise. We therefore introduce a lightweight token-filtering module that, without any attention, retains only text-relevant patches and preserves salient evidence. Because the token selection is discrete, we train it end-to-end with a straight-through estimator (STE) \cite{STE}, which allows gradients to pass through the hard selection mask.

\noindent\textbf{Scoring and STE-based Top-$K$ filtering.}
Let $k\in(0,1]$ denote the \emph{keep ratio} defined as the fraction of patches to retain. Let $K$ be the number of patches retained per frame.
We first project the question to the patch channel and compute per-patch compatibility scores:
\[
\mathbf{t}' = W_t \mathbf{t} + \mathbf{b}_t,\qquad
\mathbf{s}_t \;=\; \frac{\mathbf{t}'\,(\mathbf{v}^{\mathrm{p}}_{t})^{\!\top}}{\sqrt{D'}}\ \in \mathbb{R}^{P},
\]
We select the indices of the top-$K$ scores:
\[
I_t \;=\; \mathrm{TopK}(\mathbf{s}_t, K),
\]
Next, we gather the corresponding patch tokens
\[
X^{\mathrm{hard}}_{t} \;=\; \mathbf{v}^{\mathrm{p}}_{t}[I_t] \;\in\; \mathbb{R}^{K\times D'}.
\]
To keep the forward path identical to the hard selection while allowing a gradient flow into the scoring path, we adopt a straight-through estimator (STE) \cite{STE}. Specifically, let $\tau>0$ be a temperature and define the soft weights and their top-$K$ slice:
\[
\mathbf{w}_t \;=\; \mathrm{softmax}\!\Big(\tfrac{\mathbf{s}_t}{\tau}\Big)\in\mathbb{R}^{P},\qquad
\mathbf{w}^{\mathrm{top}}_t \;=\; \mathbf{w}_t[I_t]\in\mathbb{R}^{K}.
\]
We form a softly reweighted selection by broadcasting $\mathbf{w}^{\mathrm{top}}_t$ across feature dimensions:
\[
X^{\mathrm{soft}}_{t}
\;=\;
X^{\mathrm{hard}}_{t}\ \odot\ \mathbf{w}^{\mathrm{top}}_t\ \in\ \mathbb{R}^{K\times D'}.
\]
where \(\odot\) denotes elementwise multiplication.
Finally, we use the STE composition using stop-gradient $\text{sg}(\cdot)$:
\[
\mathbf{v}^{\mathrm{p}}_{t}
\;=\;
\mathrm{sg}\!\big(X^{\mathrm{hard}}_{t}\big)
\;+\;
\Big(
X^{\mathrm{soft}}_{t}
\;-\;
\mathrm{sg}\!\big(X^{\mathrm{soft}}_{t}\big)
\Big).
\]
Thus, the forward pass satisfies $X_t \equiv X^{\mathrm{hard}}_{t}$, whereas the backward pass routes gradients through $\mathbf{w}^{\mathrm{top}}_t$ into the scores $\mathbf{s}_t$ and ultimately into $W_t$.
In all experiments, we set the keep ratio to $k\!=\!0.5$ (i.e., $K=P/2$).

\noindent\textbf{Self-attention refinement.}
After the filtering stage, we enhance the representations by applying self-attention to each stream—the selected patch-token set \(\mathbf{V^p}\), the frame-level video \(\mathbf{V}\), the audio \(\mathbf{A}\), and the question \(\mathbf{t}\).

\subsection{Attention Framework for Modality Fusion}\label{sec:attn}

% \begin{figure}[t]
%   \centering
%   \includegraphics[width=0.75\linewidth]{figs/fig3.png} % 55%로 축소
%   \caption{CSS MATRIX}
%   \label{figure 2}
% \end{figure}

\noindent\textbf{Overview.} 
%To avoid stacked attention, we perform a wide, distribution-level fusion: the question-conditioned expectations $\mathbb{E}[\mathbf{V}|\mathbf{t}]$ and $\mathbb{E}[\mathbf{A}|\mathbf{t}]$, marginalize over the audio–video grid, yielding cross-modal evidence without deep sequential stacking. 
We propose a new framework for multi-modal attention which fuses  video and audio frame representations conditional on the input text.
The key task for AVQA is modeling the relation between video and audio given the input text (query). For that purpose, we define Conditional Similarity Score (CSS) %~\ref{figure 2} 
as follows. The CSS between $i$-th video frame and $j$-th audio frame conditional on text query $\mathbf{t}$ is denoted by $s(i,j|\mathbf{t})\in\mathbb{R}$. We define CSS matrix $\mathbf{S}\in\mathbb{R}^{T\times T}$ that has $s(i,j|\mathbf{t})$ as the $(i,j)$-th element, i.e.,
\begin{align}
    [\mathbf{S}]_{i,j} = s(i,j|\mathbf{t})\label{eq:S}
\end{align}
Thus, the row direction of $\mathbf{S}$ represents video frames, and column direction of $\mathbf{S}$ represents audio frames. % (see Figure). %It represents the similarity between $i$-th frame video and $j$-th frame audio. 
We define a probability distribution from $\mathbf{S}$ by applying a 2-D softmax to $\mathbf{S}$, i.e.,
\begin{align}
P_{\mathbf{S}}(i,j|\mathbf{t}) = \dfrac{\exp(s(i,j|\mathbf{t}))}{\sum_{i'}\sum_{j'} \exp(s(i',j'|\mathbf{t}))}
\end{align}
Thus, $P_{\mathbf{S}}(i,j|\mathbf{t})$ can be considered as a relative probability mass (weight) encoding the importance of $i$-th video and $j$-th audio frame conditional on question $\mathbf{t}$. We consider three types of CSS $s(i,j|\mathbf{t})$ as follows.

\paragraph{1) Default CSS.} Let $\mathbf{V}=\{\mathbf{v}_k\}_{k=1}^T$ and $\mathbf{A}=\{\mathbf{a}_k\}_{k=1}^T$ be $T\times D$ matrices of video and audio representations respectively. The first CSS $s(i,j|\mathbf{t})$ for $i$-th video and $j$-th audio frames is defined as
\begin{align}
s(i,j|\mathbf{t}) = \mathbf{v}_i \mathbf{t}^\top + \mathbf{a}_j \mathbf{t}^\top\label{eq:css1}
\end{align}
The CSS is a combination of the similarity between $\mathbf{v}_i$ and query $\mathbf{t}$ and the similarity between $\mathbf{a}_j$ and query $\mathbf{t}$. We derive the video embedding as the weighted sum with respect to the distribution $P_{\mathbf{S}}(i,j|\mathbf{t})$. Equivalently, it can be considered as a \emph{conditional expectation}:
\begin{align}
\mathbb{E}\left[\mathbf{V}|\mathbf{t}\right] := \sum_{i,j} P_{\mathbf{S}}(i,j|\mathbf{t})\cdot\mathbf{v}_i
\end{align}
We have that
\begin{align}
\mathbb{E}\left[\mathbf{V}|\mathbf{t}\right] = \text{Attn}(\mathbf{t},\mathbf{V},\mathbf{V})\label{eq:ca-v}
\end{align}
because
\begin{align}    &\mathbb{E}\left[\mathbf{V}|\mathbf{t}\right]=\sum_{i,j}\mathbf{v}_{i}\cdot P_{\mathbf{S}}(i,j|\mathbf{t}) 
    \\&= 
\sum_{i,j} \dfrac{\mathbf{v}_i\exp(\mathbf{v}_i \mathbf{t}^\top + \mathbf{a}_j \mathbf{t}^\top)}{\sum_{i'}\sum_{j'} \exp(\mathbf{v}_{i'} \mathbf{t}^\top + \mathbf{a}_{j'} \mathbf{t}^\top)}
\\&=\dfrac{\sum_i\mathbf{v}_i\exp(\mathbf{v}_i \mathbf{t}^\top)\cdot\sum_j\exp(\mathbf{a}_j \mathbf{t}^\top)}{\sum_{i'}\exp(\mathbf{v}_{i'} \mathbf{t}^\top)\cdot\sum_{j'}\exp(\mathbf{a}_{j'} \mathbf{t}^\top)}
\\&= \sum_i \mathbf{v}_i\cdot\dfrac{\exp(\mathbf{v}_i \mathbf{t}^\top)}{\sum_{i'}\exp(\mathbf{v}_{i'} \mathbf{t}^\top)}= \text{Attn}(\mathbf{t},\mathbf{V},\mathbf{V})\label{eq:deriv}
\end{align}
Similarly, we can show
\begin{align}
\mathbb{E}\left[\mathbf{A}|\mathbf{t}\right] = \text{Attn}(\mathbf{t},\mathbf{A},\mathbf{A})\label{eq:ca-a}
\end{align}
Thus, with CSS given by \eqref{eq:css1}, we obtain the conventional \emph{cross attention} for video and audio with input text given by \eqref{eq:ca-v} and \eqref{eq:ca-a}.

\paragraph{2) Cross-Reference CSS.} The second type of CSS is
\begin{align}
\bar s(i,j|\mathbf{t}) = \mathbf{a}_i \mathbf{t}^\top + \mathbf{v}_j \mathbf{t}^\top
\label{eq:css2}
\end{align}
Importantly, we use $i$-th \emph{audio} frame $\mathbf{a}_i$ to represent the $i$-th \emph{video} frame for similarity score. Similarly, we use $j$-th \emph{video} frame $\mathbf{v}_j$ to represent $j$-th \emph{audio} frame. The intuition is that, audio-visual questions are likely to address the \emph{co-occurrence} of video and audio events \cite{videoaudio,videoaudio2}. If the $i$-th frame contains an important clue to the audio-visual question, \emph{both} the video and audio information at the $i$-th frame will be important. In that case, the video information at $i$-th frame can be closely related to $i$-th audio $\mathbf{a}_i$ given query $\mathbf{t}$.

The CSS matrix for \eqref{eq:css2} is simply the \emph{transpose} of CSS matrix $\mathbf{S}$ in \eqref{eq:S}, because $\bar s(i,j|\mathbf{t}) = s(j,i|\mathbf{t})$. Thus, we will denote its CSS matrix by $\mathbf{S}^\top$. We obtain the softmax distribution by applying 2D-softmax to $\mathbf{S}^\top$ given by ${P_{\mathbf{S}^\top}}$. The expected representations of video and audio under this distribution can be derived similarly to \eqref{eq:deriv}:
\begin{align}
\mathbb{E}\left[\mathbf{V}|\mathbf{t}\right] = \text{Attn}(\mathbf{t},\mathbf{A},\mathbf{V})\label{eq:tma-v}
\end{align}
and
\begin{align}
\mathbb{E}\left[\mathbf{A}|\mathbf{t}\right] = \text{Attn}(\mathbf{t},\mathbf{V},\mathbf{A})\label{eq:tma-a}
\end{align}
The attentions in \eqref{eq:tma-v} and \eqref{eq:tma-a} may appear somewhat unconventional; three different modalities are input to $Q$, $K$ and $V$ connections of the attention function. However, we can interpret \eqref{eq:tma-v} as follows. The attention weights are derived from text and audio ($\mathbf{t}$ and $\mathbf{A}$ are connected to $Q$ and $K$ of the attention function). The weights ``select'' important audio frames given the question. These weights are used to combine video frames. Thus, $ \text{Attn}(\mathbf{t},\mathbf{A},\mathbf{V})$ is a combination of video frames where the frames are considered important from the perspective of audio and text. We can expect these representations are useful for audio-visual queries. We call \eqref{eq:tma-v} and \eqref{eq:tma-a} \emph{Tri-Modal} attention. %In our work, we will use four expectations of representations: \eqref{eq:ca-v} and \eqref{eq:tma-v} for video, and \eqref{eq:ca-a} and \eqref{eq:tma-a} for audio.

%suppress possible noises in the scores. For example, if we took the sum irrelevant pooling

%The first and second types of CSS can be regarded as computing aggregate cross-frame similarity between distinct frame pairs. However, we seek a unified CSS that considers both views jointly. Viewing the two types as alignments, we merge them by retaining the strongest signal—a Mixture of Alignments approach. Inspired by max pooling (which keeps the most salient activation within a region) and maxout\cite{MAXOUT} (which outputs the maximum over multiple linear responses), we adopt a hard mixture, taking the maximum between the two alignment scores for each pair $(i,j)$, i.e., we fuse by selecting the larger of the two. This is because, For example, the direct alignment $(\mathbf{v}_i,\mathbf{a}_j)$ may contain the necessary signal for the query, whereas the cross alignment $(\mathbf{a}_i,\mathbf{v}_j)$ may be unnecessary for answering the question (or vice versa). This hard selection suppresses such noise by choosing only the more plausible alignment for each pair $(i,j)$. 

\paragraph{3) Mixed CSS.} The third type of CSS we use is
\begin{align}
\hat s(i,j|\mathbf{t}) = \max(\mathbf{v}_i \mathbf{t}^\top + \mathbf{a}_j \mathbf{t}^\top,\mathbf{a}_i \mathbf{t}^\top + \mathbf{v}_j \mathbf{t}^\top)\label{eq:css3}
\end{align}
\eqref{eq:css3} can be considered as \emph{mixing} the previous CSSs \eqref{eq:css1} and \eqref{eq:css2} by taking their maximum. We chose the max function for mixing scores due the following reasons. The max function is widely adopted to combine or pool feature scores, e.g., max pooling \cite{AlexNet} (keeps the most salient activation) or maxout \cite{MAXOUT} (outputs the maximum over multiple linear responses). In particular, the max function is effective for suppressing irrelevant noise. For example, suppose one of feature score is high (large positive number) and the other is low (large negative number). Then this feature is potentially important, similar to a neuron output with a high activation value. However, if we mix the scores using sum or product, the irrelevant score can affect the result. Thus, the max operation is good at suppressing such irrelevant noise.

An issue with the max function is that it is not a smooth function. Instead, we use Log-Sum-Exp \cite{boyd2004convex} as a smooth approximation of the max function:
\begin{align}
\max(a,b)\approx \log(\exp(a) + \exp(b))\label{eq:lse}
\end{align}
If we apply \eqref{eq:lse} to CSS \eqref{eq:css3} and compute the expectations of video and audio representations, we obtain the following 
%Taking the expectation yields the expression below;
(a derivation is provided in the Supplementary Materials):
\begin{align}
\mathbb{E}[\mathbf{V}|\mathbf{t}]
\approx \tfrac12\,\text{Attn}(\mathbf{t},\mathbf{A},\mathbf{V})
+ \tfrac12\,\text{Attn}(\mathbf{t},\mathbf{V},\mathbf{V})\label{eq:mix1}
\end{align}
and 
\begin{align}
\mathbb{E}[\mathbf{A}|\mathbf{t}]
\approx \tfrac12\,\text{Attn}(\mathbf{t},\mathbf{V},\mathbf{A})
+ \tfrac12\,\text{Attn}(\mathbf{t},\mathbf{A},\mathbf{A}).\label{eq:mix2}
\end{align}
Interestingly, these approximate representations can be obtained from the \emph{mixture} of softmax distributions derived from previous CSSs. For example, \eqref{eq:mix1} is equal to
\[
\sum_{i,j} \left( \frac{P_{\mathbf{S}}(i,j|\mathbf{t}) +  P_{\mathbf{S}^\top}(i,j|\mathbf{t})}{2}\right) \mathbf{v}_i
\]
which means that \eqref{eq:mix1} is the expectation with respect to the mixture of distributions $P_{\mathbf{S}}$ and $P_{\mathbf{S}^\top}$.
%This expression coincides with taking the results obtained in second type of CSS, third type of CSS and averaging them with equal weights \(1/2\).

\noindent \textbf{Implementation: Modality Mixture.}
We obtained representations which are cross attentions ($ \text{Attn}(\mathbf{t},\mathbf{V},\mathbf{V})$ and
$ \text{Attn}(\mathbf{t},\mathbf{A},\mathbf{A})$ in \eqref{eq:ca-v} and \eqref{eq:ca-a}), Tri-Modal Attentions ($\text{Attn}(\mathbf{t},\mathbf{A},\mathbf{V})$ and $ \text{Attn}(\mathbf{t},\mathbf{V},\mathbf{A})$ in \eqref{eq:tma-v} and \eqref{eq:tma-a}), and their mixtures (\eqref{eq:mix1} and \eqref{eq:mix2}). Thus, these attentions constitute the main components of the proposed framework for modality fusion. For practical implementation, we  combine these representations as follows: (i) compute the mixture weights for the four attention representations; (ii) concatenate the weighted attention representations. The concatenated representation will be used for the final reasoning of the answer. This approach is consistent with our horizontal fusion of modalities; we compute four types of attentions directly from audio, video and text representations, and merge them in parallel without deep sequential stacking.

%To avoid stacked attention, we perform a wide, distribution-level fusion: the question-conditioned expectations $\mathbb{E}[\mathbf{V}|\mathbf{t}]$ and $\mathbb{E}[\mathbf{A}|\mathbf{t}]$, marginalize over the audio–video grid, yielding cross-modal evidence 
%For the $V$, we use both frame-level and patch-level embeddings: $\operatorname{Attn}(\mathbf{t}, \mathbf{V}, \mathbf{A})$ operates on frame-level features, while all other attention modules use patch-level embeddings to better exploit spatial information.
Denote these cross- and Tri-Modal attention outputs by $ r_1= \text{Attn}(\mathbf{t},\mathbf{V}^p,\mathbf{V}^p)$, 
$r_2=\text{Attn}(\mathbf{t},\mathbf{A},\mathbf{A})$, $r_3=\text{Attn}(\mathbf{t},\mathbf{A},\mathbf{V}^p)$ and $ r_4=\text{Attn}(\mathbf{t},\mathbf{V},\mathbf{A})$, respectively. Note that we used patch-level representations of video $(\mathbf{V}^p)$ for the Value input to the attentions to better exploit spatial information.
\color{black}
The attention outputs are fused using text-dependent weights. A precise weighting scheme is described as follows. For stability of those features, we modulate attention keys and values by text using FiLM\cite{FILM}:
\begin{align*}
\mathrm{FiLM}(X;\mathbf{t}) &\;=\; X \odot \gamma(\mathbf{t}) + \beta(\mathbf{t})
\\\gamma(\mathbf{t})&=\mathbf{1}+W_\gamma \mathbf{t},\;\; \beta(t)=W_\beta \mathbf{t},
\end{align*}
(zero-initialized so initially identity).
For $r_1$, $r_2$, $r_3$, and $r_4$, we use
\[
 K'=\mathrm{FiLM}(K;\mathbf{t}),\qquad V'=\mathrm{FiLM}(V;\mathbf{t}),
\]
and apply attention with $(K',V')$ in place of $(K,V)$ inside Attention Module.
From $t$ we produce four scalars
\[
(q_1,q_2,q_3,q_4)=W_g \mathbf{t}+ b_g,\qquad 
g_i \;=\; 1+\tfrac{1}{2}\tanh(q_i).
\]
Each attention is gated by $g_i$:
\[
h_i \;=\; g_i\, r_i,\qquad i=1,\dots,4.
\]
This approach allows the question \(\mathbf{t}\) to determine how much each attention result \(r_i\) should matter by assigning weights \(g_i\) that control their contribution.
% \color{red}
% In doing so, we realize \emph{parallel} modality fusion by computing and mixing all interaction candidates in a single stage, while Tri-Modal Attention explicitly enables direct three-way interaction among text, video, and audio for AVQA reasoning.
% \color{black}

\subsection{Reasoning}
For the reasoning of the answer, we begin with concatenating the text embedding and four gated attention representations:
\[\mathbf{H} \;=\; \big[\mathbf{t}\;;\;h_1\;;\;h_2\;;\;h_3\;;\;h_4\,\big] \in \mathbb{R}^{5\times D}\]
Next, we apply a self-attention block to allow global interaction among all results:
\[\mathbf{F} \;=\; \mathrm{SA}\!\left(\mathbf{H},\,\mathbf{H},\,\mathbf{H}\right) \in \mathbb{R}^{5\times D}
\]
Finally, the result passes through a classification layer to produce the answer:
\[
\hat{\mathbf{y}} \;=\; \text{CLS}(\mathbf{F}) \;\in\; \mathbb{R}^{C}
\]
where $\text{CLS}$ denotes the classifier layer and $C$ denotes the number of classes.
%This stage turns the pathway-specific evidence into a single, contextually mixed representation from which the decision is made. 
% \begin{align*}
% \mathbf{H} &\;=\; \big[Q\;;\;h_1\;;\;h_2\;;\;h_3\;;\;h_4\,\big] \in \mathbb{R}^{5\times D},
% \\\mathbf{F} &\;=\; \mathrm{SA}\!\left(\mathbf{H},\,\mathbf{H},\,\mathbf{H}\right) \in \mathbb{R}^{5\times D}.
% \\\hat{\mathbf{y}} & \;=\; \mathrm{CLS}(\mathbf{F}) \;\in\; \mathbb{R}^{C}
% \end{align*}
The model is trained by minimizing the cross-entropy between the predicted label $\hat{\mathbf{y}}$ and the ground-truth label $\mathbf{y}$:
\begin{equation}
\mathcal{L}_{\mathrm{CE}} = - \sum_{c=1}^{C} y_c \log \hat{y}_c \, .
\end{equation}

\section{Experiments}

\subsection{Datasets}
We evaluate on MUSIC-AVQA \cite{MUSIC-AVQA}, MUSIC-AVQA-R \cite{MUSIC-AVQA-R}, and MUSIC-AVQA-v2.0 \cite{MUSIC-AVQA-2.0}, adopting the official train/val/test splits as in Table~\ref{table1}. The base MUSIC-AVQA \cite{MUSIC-AVQA} provides a standard benchmark over 22 instruments with audio-only, visual-only, and audio-visual questions. MUSIC-AVQA-R \cite{MUSIC-AVQA-R} builds on MUSIC-AVQA and upgrades the setting by concentrating on rare and out-of-distribution cases (e.g., uncommon instruments and atypical contexts), explicitly probing robustness beyond head-class priors and frequency-driven shortcuts. MUSIC-AVQA-v2.0 \cite{MUSIC-AVQA-2.0} upgrades the benchmark to address dataset biases by enriching ensemble and multi-instrument scenes, rebalancing question types (e.g., existence, location, temporal) across modalities. It provides paired bias-balanced evaluation splits, which enable controlled analysis of spurious correlations versus genuine cross-modal reasoning.

\begin{table}[t]
\caption{Dataset statistics.}
\centering
{\scriptsize
\setlength{\tabcolsep}{3pt}      % 칸 간격
\renewcommand{\arraystretch}{1.2} % 행 간격
\begin{tabular}{l|c|c|c|c}
\Xhline{0.8pt}
\textbf{Dataset} & \textbf{Videos} & \textbf{Train} & \textbf{Valid QA} & \textbf{Test QA} \\
\hline
MUSIC-AVQA        & 9,288  & 31,927 & 4,568 & 9,129   \\
MUSIC-AVQA-R      & 9,288  & --     & --    & 211,572 \\
MUSIC-AVQA-v2.0   & 10,512 & 37,429 & 5,345 & 10,816  \\
\Xhline{0.8pt}
\end{tabular}}
\label{table1}
\end{table}

\subsection{Implementation Details}
We extract common-space representations by using CLIP \cite{CLIP} for video and text embeddings, and ImageBind \cite{imagebind} for audio embeddings. Video frames are sampled every 2 seconds with the hidden size given by \(D=1024\), and we obtain one CLIP embedding per frame. However, this frame-wise CLIP encoding may not fully represent spatial information. Thus, we additionally extract CLIP-based patch-level embeddings for video: for each frame, \(P=256\) patch tokens with dimensionality \(1280\).
All attention modules use 8 heads. Dropout 0.1 is applied to every attention module and to the final classification layer. Optimization uses AdamW \cite{adamw} with a step scheduler that multiplies the learning rate by \({0.1}\) every 8 epochs. The loss employs label smoothing \cite{labelsmoothing} \(0.1\). We apply gradient clipping of \(0.1\) (global norm). Training is performed on a single NVIDIA A6000 GPU.

\subsection{Quantitative Results and Analysis}
\label{sec:results}

\noindent\textbf{MUSIC-AVQA-R.}
On the debiased MUSIC-AVQA-R\cite{MUSIC-AVQA-R} benchmark, our model attains 70.89\% overall as in Table~\ref{table2}. It performs strongly on both frequent (Head) and rare (Tail) answer categories, e.g., Audio QA Count (H/T: 85.68/76.12\%) and Visual QA Local (H/T: 89.45/78.95\%). For Audio-Visual QA, Count (H/T: 81.43/44.21\%) and Local (H/T: 66.27/63.51\%) remain competitive, suggesting robust question-conditioned temporal–spatial reasoning even under long-tail distributions.

\noindent\textbf{MUSIC-AVQA.}
Our model achieves an overall accuracy of 77.68\% on the MUSIC-AVQA\cite{MUSIC-AVQA} test set, surpassing prior work such as QA-TIGER (77.62\%) as shown in Table~\ref{table3}. By modality, we obtain a strong Visual QA average of 85.76\% (Count 84.46\%, Local 87.02\%) and an Audio-Visual QA average of 73.78\% (Exist 82.29\%, Count 80.87\%, Local 71.85\%, Comp 62.58\%, Temp 69.83\%). Audio QA averages 77.84\%. 

\noindent\textbf{MUSIC-AVQA-v2.0.}
We further evaluate on the MUSIC-AVQA-v2.0\cite{MUSIC-AVQA-2.0} benchmark. When trained on the biased dataset, Q-TriM achieves state-of-the-art performance on both the bias test (78.40\%) and the balanced test (75.31\%). When trained on the balanced dataset, it remains state-of-the-art performance on the bias test (76.92\%), but on the balanced test it falls short by a small margin (76.29\%), as shown in Table~\ref{table4}.

Overall, Q-TriM attains strong results across all three benchmarks. %On MUSIC-AVQA-R it surpasses the previous state of the art, QA-TIGER \cite{QATIGER} by 2.9\%. 
This suggests that adopting a shallow and parallel design, rather than deep and sequential, for inter-modal attention can be effective for learning cross-modal interactions compared to deeper stacked architectures as in baseline methods.

\begin{table}[t]
\caption{Experimental results (\%) on the MUSIC-AVQA-R test set, with H and T representing performance on Head (frequent) and Tail (rare) answer categories, respectively. \textbf{Bold} denotes the best performance, and \underline{underline} indicates the second-best performance.}
\label{table2}
\centering

% Method | Audio QA(Count H/T, Comp H/T) | Visual QA(Count H/T, Local H/T)
% | Audio-Visual QA(Exist/Count/Local/Comp/Temp : 각 H/T) | Avg
\setlength{\tabcolsep}{2pt}     % 열 간격 축소 (기존 4.5pt → 2.4pt)
\renewcommand{\arraystretch}{1.2}% 행간 약간만 줄임 (가독성 유지)
\resizebox{\textwidth}{!}{
% H/T 라벨 더 줄이기
\newcommand{\HTH}{\tiny H}
\newcommand{\HTT}{\tiny T}
\begin{tabular}{l|cc|cc|cc|cc|cc|cc|cc|cc|cc|c}
\hline
\multicolumn{1}{c|}{\multirow{3}{*}[-0.6ex]{\textbf{Method}}} &
\multicolumn{4}{c|}{\textbf{Audio QA}} &
\multicolumn{4}{c|}{\textbf{Visual QA}} &
\multicolumn{10}{c|}{\textbf{Audio-Visual QA}} &
\multicolumn{1}{c}{\textbf{Avg}} \\
\cline{2-5}\cline{6-9}\cline{10-19}
& \multicolumn{2}{c|}{Count} & \multicolumn{2}{c|}{Comp} &
  \multicolumn{2}{c|}{Count} & \multicolumn{2}{c|}{Local} &
  \multicolumn{2}{c|}{Exist} & \multicolumn{2}{c|}{Count} &
  \multicolumn{2}{c|}{Local} & \multicolumn{2}{c|}{Comp} &
  \multicolumn{2}{c|}{Temp} &  \\ % ← Avg는 H/T 없음!
\cline{2-3}\cline{4-5}\cline{6-7}\cline{8-9}\cline{10-11}\cline{12-13}\cline{14-15}\cline{16-17}\cline{18-19}
& H & T & H & T & H & T & H & T & H & T & H & T & H & T & H & T & H & T & \\

\hline
FCNLSTM\cite{FCNLSTM}   & 66.23 & 36.48 & 64.78 & 51.24 & 61.75 & 5.31 & 54.86 & 51.06 & 64.76 & 78.52 & 62.69 & 7.23 & 46.66 & 57.30 & 43.13 & 71.67 & 37.02 & 30.78 & 54.12 \\
BiLSTM\cite{BILSTM}    & 73.68 & 46.32 & 21.51 & \textbf{77.58} & 64.30 & 0.00 & 53.92 & 42.01 & \textbf{87.51} & 21.14 & 62.85 & 2.18 & 35.16 & 43.75 & 27.61 & \underline{74.38} & 17.58 & 31.32 & 48.84 \\
HCAttn\cite{HCAttn}    & 61.67 & 41.63 & 59.09 & 47.14 & 56.52 & 9.20 & 67.01 & 53.16 & 66.57 & 61.13 & 59.53 & 12.48 & 37.05 & 42.48 & 48.81 & 60.12 & 33.82 & 39.26 & 51.90 \\
MCAN\cite{MCAN}      & 75.02 & 60.16 & 58.89 & 50.90 & 64.58 & 26.69 & 66.62 & 65.25 & 61.29 & 67.29 & 64.76 & 25.28 & 46.11 & 61.61 & 50.57 & 52.04 & 34.64 & 58.05 & 57.27 \\
PSAC\cite{PSAC}      & 53.01 & 56.68 & 57.41 & 48.12 & 49.55 & 26.43 & 72.96 & 60.69 & 50.56 & 55.54 & 56.70 & 19.58 & 41.98 & 52.30 & 38.13 & 58.92 & 26.68 & 46.24 & 50.45 \\
HME\cite{HME}       & 62.60 & 53.95 & 54.97 & 58.29 & 50.95 & 16.46 & 73.25 & 58.06 & 65.74 & 66.49 & 63.18 & 17.18 & 39.59 & 45.03 & 53.20 & 69.57 & 39.35 & 41.57 & 53.66 \\
HCRN\cite{HCRN}      & 55.53 & 53.31 & 47.17 & 32.44 & 41.87 & 23.55 & 39.40 & 51.27 & 41.81 & 65.45 & 54.58 & 19.57 & 36.62 & 42.72 & 33.33 & 36.87 & 40.47 & 44.13 & 43.92 \\
AVSD\cite{AVSD}      & 54.00 & 47.84 & 60.61 & 47.79 & 63.25 & 10.07 & 74.70 & 61.43 & 66.28 & 61.98 & 42.61 & 18.06 & 33.00 & 40.35 & 51.98 & 66.00 & 40.14 & 41.52 & 52.33 \\
Pano-AVQA\cite{PANO-AVQA} & 50.57 & 43.45 & 50.78 & 43.64 & 47.28 & 15.50 & 76.19 & 65.51 & 52.37 & 22.04 & 52.21 & 21.52 & 44.35 & \underline{61.69} & 45.61 & 40.49 & 40.05 & 39.43 & 47.40 \\
ST-AVQA\cite{MUSIC-AVQA}   & 56.40 & 41.48 & 62.22 & 55.79 & 59.86 & 12.94 & 63.31 & 54.00 & 49.78 & 56.41 & 38.11 & 19.41 & 35.05 & 44.09 & 53.30 & 62.44 & 40.42 & 35.85 & 52.80 \\
LAVISH\cite{LAVISH}    & 61.73 & 43.99 & 66.00 & 50.38 & 55.33 & 10.21 & 70.71 & 68.78 & \underline{77.83} & 79.46 & 49.38 & 14.47 & 41.76 & 41.01 & 59.26 & 65.10 & 41.84 & 46.26 & 57.63 \\
TSPM\cite{TSPM}      & 81.65 & 71.80 & \underline{67.66} & 49.56 & 78.29 & 47.56 & 58.80 & 73.18 & 69.15 & \textbf{82.79} & \underline{77.09} & 38.64 & 42.24 & 57.37 & 52.07 & 68.86 & 39.23 & 49.36 & 66.30 \\
MCCD\cite{MUSIC-AVQA-R}      & \underline{84.32} & 67.23 & 64.68 & 62.18 & 75.09 & 48.42 & 80.47 & 66.38 & 77.22 & 67.58 & 55.15 & \textbf{82.23} & \textbf{70.12} & 39.83 & \underline{61.26} & 58.17 & \underline{43.67} & \underline{58.33} & 66.95 \\
QA-Tiger\cite{QATIGER}  & 82.67 & \underline{75.82} & \textbf{71.75} & 43.11 & \underline{81.30} & \underline{54.59} & \underline{84.76} & \underline{75.59} & 72.84 & 78.56 & 76.70 & 33.55 & 64.65 & \textbf{64.65} & 37.55 & \textbf{80.47} & 36.85 & \textbf{62.96} & \underline{67.99} \\
\hline
\textbf{Q-TriM} & \textbf{85.68} & \textbf{76.12} & 55.22 & \underline{63.54} & \textbf{86.26} & \textbf{57.91} & \textbf{89.45} & \textbf{78.95} & 72.29 & \underline{81.97} & \textbf{81.43} & \underline{44.21} & \underline{66.27} & 61.35 & \textbf{61.35} & 56.67 & \textbf{43.94} & 48.26 & \textbf{70.89} \\
\hline
\end{tabular}

}

\bigskip

\caption{Experimental results (\%) on the MUSIC-AVQA test set.}
\centering
\setlength{\tabcolsep}{7pt}
\renewcommand{\arraystretch}{1.0}
\resizebox{\textwidth}{!}{
\begin{tabular}{@{}l|ccc|ccc|cccccc|c@{}}
\hline
\multicolumn{1}{c|}{\multirow{3}{*}[1ex]{\textbf{Method}}}&
\multicolumn{3}{c|}{\textbf{Audio QA}} &
\multicolumn{3}{c|}{\textbf{Visual QA}} &
\multicolumn{6}{c|}{\textbf{Audio-Visual QA}} &
\multicolumn{1}{c}{\textbf{Avg}} \\
\cline{2-4}\cline{5-7}\cline{8-13}
& Count & Comp & Avg & Count & Local & Avg & Exist & Count & Local & Comp & Temp & Avg & \\
\hline
FCNLSTM\cite{FCNLSTM}     & 70.45 & 66.22 & 68.88 & 63.89 & 46.74 & 55.21 & 82.01 & 59.34 & 46.28 & 62.15 & 47.33 & 60.06 & 60.34 \\
BiLSTM\cite{BILSTM}     & 70.35 & 47.92 & 62.05 & 64.64 & 64.33 & 64.48 & 78.39 & 56.91 & 45.85 & 53.09 & 49.76 & 57.10 & 59.92 \\
HCAttn\cite{HCAttn}     & 70.25 & 54.91 & 64.57 & 64.05 & 66.37 & 65.22 & 79.10 & 59.97 & 49.51 & 55.25 & 56.43 & 60.19 & 62.30 \\
MCAN\cite{MCAN}        & 77.50 & 55.24 & 69.25 & 71.56 & 70.93 & 71.24 & 80.40 & 64.91 & 54.48 & 57.22 & 47.57 & 61.58 & 65.49 \\
PSAC\cite{PSAC}        & 75.64 & 66.06 & 72.09 & 68.64 & 69.79 & 69.22 & 77.59 & 63.42 & 55.02 & 61.17 & 59.47 & 63.52 & 66.54 \\
HME\cite{HME}         & 74.76 & 63.56 & 70.61 & 67.97 & 69.46 & 68.76 & 80.30 & 63.19 & 53.18 & 62.69 & 59.83 & 64.05 & 66.45 \\
HCRN\cite{HCRN}       & 68.59 & 50.92 & 62.05 & 64.39 & 61.81 & 63.08 & 54.47 & 53.38 & 41.53 & 52.11 & 47.69 & 50.26 & 55.73 \\
AVSD\cite{AVSD}        & 72.41 & 61.90 & 68.52 & 67.39 & 74.19 & 70.83 & 81.61 & 63.89 & 58.79 & 61.52 & 61.41 & 65.49 & 67.44 \\
Pano-AVQA\cite{PANO-AVQA}  & 74.36 & 64.56 & 70.73 & 69.39 & 75.65 & 72.56 & 81.21 & 64.91 & 59.33 & 64.22 & 63.23 & 66.64 & 68.93 \\
ST-AVQA\cite{MUSIC-AVQA}     & 78.18 & 67.05 & 74.06 & 71.56 & 76.38 & 74.00 & 81.81 & 70.80 & 64.51 & 66.01 & 63.23 & 69.54 & 71.52 \\
MCCD\cite{MUSIC-AVQA-R}       & 83.87 & \textbf{71.04} & \textbf{79.14} & 79.78 & 76.73 & 78.24 & 80.87 & 51.63 & 71.46 & 64.67 & 64.60 & 67.13 & 72.20 \\
COCA\cite{COCA}      & 79.35 & 67.68 & 75.42 & 75.10 & 75.43 & 75.23 & \textbf{83.50} & 66.63 & 69.72 & 64.12 & 65.57 & 69.96 & 72.33 \\
PSTP-Net\cite{PSTP-NET}   & 73.97 & 65.59 & 70.91 & 77.15 & 77.36 & 77.26 & 76.18 & 72.23 & 71.80 & \textbf{71.79} & 69.00 & 72.57 & 73.52 \\
LAVISH\cite{LAVISH}     & 82.09 & 65.56 & 75.97 & 78.98 & 81.43 & 80.22 & 81.71 & 75.51 & 66.13 & 63.77 & 67.96 & 71.26 & 74.46 \\
APL\cite{APL}        & 82.40 & \underline{70.71} & 78.09 & 76.52 & 82.74 & 79.69 & 82.99 & 73.29 & 66.68 & 64.76 & 65.95 & 70.96 & 74.53 \\
TSPM\cite{TSPM}       & 84.07 & 64.65 & 76.91 & 82.29 & 84.90 & 83.61 & 82.19 & 76.21 & \underline{71.85} & \underline{65.76} & \textbf{71.17} & 73.51 & 76.79 \\
QA-Tiger\cite{QATIGER}    & \underline{84.86} & 67.85 & \underline{78.58} & \underline{83.96} & \underline{86.29} & \underline{85.14} & \underline{83.10} & \underline{78.58} & \textbf{72.50} & 63.94 & 69.59 & \underline{73.74} & \underline{77.62} \\

\hline
\textbf{Q-TriM}     & \textbf{85.05} & 65.49 & 77.84 & \textbf{84.46} & \textbf{87.02} & \textbf{85.76} & 82.29 & \textbf{80.87} & \underline{71.85} & 62.58 & \underline{69.83} & \textbf{73.78} & \textbf{77.68} \\
\hline
\end{tabular}
}
\label{table3}

\end{table}

\begin{table}[t]
\centering
\caption{Experimental results (\%) on the MUSIC-AVQA-v2.0 for (a) bias and (b) balanced test sets. Each evaluation is conducted with models trained on bias and balanced training sets, respectively.}
\label{table4}

\setlength{\tabcolsep}{1.0pt}
\renewcommand{\arraystretch}{1.2}
\scriptsize

% (a) Bias test set
\begin{minipage}[t]{0.48\linewidth}
\centering
\resizebox{\linewidth}{!}{%
\begin{tabular}{@{}c|c|cccc@{}}
\hline
\textbf{Training} & \textbf{Method} & \textbf{A-QA} & \textbf{V-QA} & \textbf{AV-QA} & \textbf{Avg} \\
\hline
\multirow{3}{*}{Bias}
 & ST-AVQA\cite{MUSIC-AVQA}  & 76.86 & 77.70 & 69.59 & 73.07 \\
 & LAVISH\cite{LAVISH}      & 76.73 & 80.96 & 70.80 & 74.59 \\
 & QA-TIGER\cite{QATIGER}   & \textbf{79.13} & \underline{84.83} & \underline{72.37} & \underline{76.93} \\
\cline{2-6}
 & \textbf{Ours}            & \underline{78.93} & \textbf{86.01} & \textbf{74.53} & \textbf{78.40} \\
\cline{1-6}
\multirow{5}{*}{Balance}
 & ST-AVQA\cite{MUSIC-AVQA} & 76.18 & 77.20 & 67.96 & 71.92 \\
 & LAVISH\cite{LAVISH}      & 75.56 & 80.83 & 69.27 & 73.51 \\
 & LAST\cite{MUSIC-AVQA-2.0}      & 77.10 & 82.99 & 70.86 & 75.24 \\
 & LAST-Att\cite{MUSIC-AVQA-2.0}  & \underline{77.29} & 83.47 & 71.05 & 75.45 \\
 & QA-TIGER                 & 77.07 & \underline{85.93} & \underline{71.20} & \underline{76.57} \\
\cline{2-6}
 & \textbf{Ours}            & \textbf{78.75} & \textbf{86.45} & \textbf{71.69} & \textbf{76.92} \\
\hline
\end{tabular}}
\vspace{2pt}

{\scriptsize (a) Bias test set}
\end{minipage}
\hfill
% (b) Balance test set
\begin{minipage}[t]{0.48\linewidth}
\centering
\resizebox{\linewidth}{!}{%
\begin{tabular}{@{}c|c|cccc@{}}
\hline
\textbf{Training} & \textbf{Method} & \textbf{A-QA} & \textbf{V-QA} & \textbf{AV-QA} & \textbf{Avg} \\
\hline
\multirow{3}{*}{Bias}
 & ST-AVQA\cite{MUSIC-AVQA}  & 73.34 & 76.82 & 64.51 & 69.40 \\
 & LAVISH\cite{LAVISH}       & 73.14 & 79.70 & 65.01 & 70.39 \\
 & QA-TIGER\cite{QATIGER}    & \underline{77.57} & \underline{84.84} & \underline{67.43} & \underline{73.91} \\
\cline{2-6}
 & \textbf{Ours}             & \textbf{77.67} & \textbf{85.05} & \textbf{69.85} & \textbf{75.31} \\
\cline{1-6}
\multirow{5}{*}{Balance}
 & ST-AVQA\cite{MUSIC-AVQA}  & 75.50 & 77.67 & 66.32 & 71.02 \\
 & LAVISH\cite{LAVISH}       & 76.15 & 81.32 & 68.28 & 73.18 \\
 & LAST\cite{MUSIC-AVQA-2.0} & 78.08 & 83.29 & 69.72 & 74.85 \\
 & LAST-Att\cite{MUSIC-AVQA-2.0} & 78.56 & 84.07 & \textbf{70.30} & 75.44 \\
 & QA-TIGER\cite{QATIGER}    & 79.90 & \textbf{86.95} & \underline{70.22} & \textbf{76.43} \\
\cline{2-6}
 & \textbf{Ours}             & \textbf{80.14} & \underline{86.63} & 70.03 & \underline{76.29} \\
\hline
\end{tabular}}
\vspace{2pt}

{\scriptsize (b) Balance test set}
\end{minipage}

\end{table}

\subsection{Ablation Study}\label{abl}

We conduct ablations to assess the contribution of each component of the proposed model (Table \ref{table5}).

\noindent\textbf{Token filtering.}
We first removed the token filtering component. Token filtering not only reduces computational cost, but also suppresses patches that are irrelevant to the question, thereby easing question grounding and answer selection. Without this module, the model attends more to background regions and exhibits degraded accuracy.

\noindent\textbf{Tri-Modal attention.}
Next, we removed the Tri-Modal attention. This module is particularly important for audio-visual question types in AVQA that require reasoning over all three modalities. Eliminating it forces the model to rely on cross attention layers only, which weakens cross-modal alignment and lowers performance. %Also, we replaced Tri-modal attention modules with \textbf{sequential} attention layers (4th row, Table \ref{table5}). T

\noindent\textbf{Modality mixture.}
Next, we removed the modality mixture component. This module aggregates the text-conditioned attention outputs from different modalities into a coherent representation. Without this component, the model struggles to reconcile complementary cues across modalities, leading to a further drop in accuracy.

% \paragraph{Conclusion.} Across all three ablations, performance decreases compared to the full model, confirming that each module contributes meaningfully to the overall effectiveness.

\begin{table}[t]
\centering
\caption{Ablation study of the proposed framework on MUSIC-AVQA-R dataset.}
\label{table5}
\setlength{\tabcolsep}{0.5pt} 
\renewcommand{\arraystretch}{1.0}

{\footnotesize
\begin{tabular}{c|cccc}
\Xhline{0.8pt}
\makecell{Method} &
A-QA & V-QA & AV-QA & Avg \\
\hline
\textit{w\slash o} All Modules(Base) & 70.48 & 80.83 & 58.43 & 66.96 \\
\textit{w\slash o} Token Filtering &  74.54 & 80.36 & 61.54 & 69.21 \\
\textit{w\slash o} Tri-Modal Attention & 73.13 & 81.02 & 61.10 & 68.92 \\
\textit{w\slash o} Modality Mixture & 72.08 & 78.21 & 60.96 & 67.85 \\
Ours & \textbf{74.74} & \textbf{83.64} & \textbf{62.85} & \textbf{70.89} \\
\Xhline{0.8pt}
\end{tabular}
}
\end{table}

\subsection{Keep ratio of Token Filtering}
We conducted token filtering experiments to determine how many patches to retain, controlled by the hyperparameter $k$ (Table \ref{table6}). Using the \textit{MUSIC-AVQA-R} dataset, we evaluated five values of the keep ratio: \(0.1, 0.3, 0.5, 0.7,\) and \(0.9\). Here, a keep ratio of \(0.5\) means that only \(50\%\) of the total patches are kept while the remaining ones are discarded. Consistent with the paper’s findings, a keep ratio of \(0.5\) yielded the best performance. Notably, when the keep ratio was set to \(0.1\). Thus, by using only \(10\%\) of the original patches, the model still achieved a reasonable \(70.11\%\) accuracy, indicating that token filtering effectively compresses patches. 
This suggests that, even without using attention in the filtering stage, the token-filtering module is effectively learned via simple scoring and a straight-through estimator (STE) \cite{STE} in order to retain only the question-relevant tokens.
%We also explored an adaptive keep-ratio strategy, where the model dynamically selects how many video patches to retain conditioned on the question. However, in our experiments, a fixed keep ratio consistently performed slightly better. Further details are provided in the appendix.
In addition, we explored a strategy that makes the keep-ratio adaptive, where the model dynamically selects how many video patches to retain conditioned on the question. Related discussion and experiments are provided in the Supplementary Materials. %However, in our experiments, a fixed keep ratio consistently performed slightly better. Further details are provided in the appendix.
\color{black}

\begin{table}[t]
\setlength{\tabcolsep}{0.5pt} 
\renewcommand{\arraystretch}{1}
\caption{Evaluation of token filtering with varying keep ratios on MUSIC-AVQA-R. Consistent with prior findings, a keep ratio of \(0.5\) achieves the highest accuracy.}
\centering
{\footnotesize
\begin{tabular}[0.2\linewidth]{c|cccc}
\Xhline{0.8pt}
\makecell{Keep Ratio} &
A-QA & V-QA & AV-QA & Avg \\
\hline
0.1 & 75.28 & 82.60 & 61.79 & 70.11 \\
0.3 &  75.30 & 83.05 & 62.37 & 70.56 \\
0.5(ours) & 74.74 & \textbf{83.64} & \textbf{62.85} & \textbf{70.89} \\
0.7 & 74.28 & 82.71 & 62.56 & 70.39 \\
0.9 & \textbf{75.35} & 83.22 & 62.15 & 70.50 \\
\Xhline{0.8pt}
\end{tabular}}
\label{table6}
\end{table}

\section{Conclusion}

We introduce Q-TriM, an AVQA model that replaces deeply stacked attention with a shallow, parallel inter-modal and question-aware fusion. 
% \color{red}
% effectively mitigating stage-wise information loss and preserving inter-modal cues for question-aligned audio–visual reasoning.
% \color{black}
The pipeline consists of three stages.
First, we apply token filtering to retain only the semantically meaningful patches that are relevant to the given question.
Next, Tri-Modal Attention and cross attention are applied in parallel to obtain text-relevant video and audio representations.
Finally, a modality mixture head aggregates features with question-aware weights, yielding a compact representation that emphasizes information most pertinent to the query. Q-TriM attains state-of-the-art performance on three benchmark datasets. Especially, we demonstrate the robustness of our model by achieving a substantial performance margin over prior state-of-the-art methods on the MUSIC-AVQA-R dataset. As a direction for future work, we plan to explore various CSS scores in the proposed attention framework and extend them to various multi-modal tasks.

% \clearpage  
% TODO FINAL: This \clearpage needs to be removed from both review and camera-ready versions.

\section*{Acknowledgements}
This work was partly supported by the Institute of Information \& Communications Technology Planning \& Evaluation (IITP) grant funded by the Korea government (MSIT) (IITP-2026-RS-2020-II201819, ICT Creative Consilience Program, 10\%, and No.RS-2026-25507543, Development of AI Co-Scientist based on Scientific Causal World Model, 80\%) and by the National Research Foundation of Korea (NRF) grant funded by the Korea government(MSIT)(RS-2022-NR070834,RS-2026-25482946).


\bibliographystyle{splncs04}
\bibliography{main}

% =========================================================
% Supplementary Material starts here
% =========================================================

\clearpage
\setcounter{page}{1}

\begin{center}
    {\large\bfseries Q-TriM: Question-Guided Tri-Modal Attention for\\[0.1em]
    Audio-Visual Question Answering\par}
    \vspace{0.8em}
    {\large Supplementary Material\par}
\end{center}

\vspace{0.5em}

\renewcommand{\thesection}{\Alph{section}}

\section{Derivation of Eq. (17) and Eq. (18)}
We derive Eq. (17) and Eq. (18) by applying the Log-Sum-Exp approximation in Eq. (16) to the mixed CSS given in Eq. (15). We first derive the approximation for $\mathbb{E}[\mathbf{V}\mid\mathbf{t}]$ as follows.
\begin{align}
\mathbb{E}\left[\mathbf{V}\mid\mathbf{t}\right]
&= \sum_{i,j}\mathbf{v}_{i}\cdot P_{\mathbf{S}}(i,j\mid\mathbf{t}) \\
&\approx \sum_{i,j} \mathbf{v}_i\,
   \frac{e^{\,\mathbf{v}_i \mathbf{t}^\top} e^{\,\mathbf{a}_j \mathbf{t}^\top} + e^{\,\mathbf{a}_i \mathbf{t}^\top} e^{\,\mathbf{v}_j \mathbf{t}^\top}}
        {\sum_{i',j'} \left(e^{\,\mathbf{v}_{i'} \mathbf{t}^\top} e^{\,\mathbf{a}_{j'} \mathbf{t}^\top} + e^{\,\mathbf{a}_{i'} \mathbf{t}^\top} e^{\,\mathbf{v}_{j'} \mathbf{t}^\top}\right)} \\
&=
\frac{
(\sum_i \mathbf{v}_i e^{\,\mathbf{v}_i \mathbf{t}^\top})(\sum_j e^{\,\mathbf{a}_j \mathbf{t}^\top})
+
(\sum_i \mathbf{v}_i e^{\,\mathbf{a}_i \mathbf{t}^\top})(\sum_j e^{\,\mathbf{v}_j \mathbf{t}^\top})
}{
(\sum_{i'} e^{\,\mathbf{v}_{i'} \mathbf{t}^\top})(\sum_{j'} e^{\,\mathbf{a}_{j'} \mathbf{t}^\top})
+
(\sum_{i'} e^{\,\mathbf{a}_{i'} \mathbf{t}^\top})(\sum_{j'} e^{\,\mathbf{v}_{j'} \mathbf{t}^\top})
}
\label{eq:supp}
\end{align}
The two terms in the denominator of Eq. (22) are identical products. Thus, the denominator becomes
\begin{align}
2\sum_{i'} e^{\,\mathbf{v}_{i'} \mathbf{t}^\top}\cdot\sum_{j'} e^{\,\mathbf{a}_{j'} \mathbf{t}^\top}.
\label{eq:supp2}
\end{align}
Substituting Eq. (23) into Eq. (22) yields
\begin{align}
\mathbb{E}[\mathbf{V}\mid\mathbf{t}]
&\approx \tfrac12\,
\frac{\sum_i \mathbf{v}_{i} \exp(\mathbf{v}_i \mathbf{t}^\top)}
     {\sum_{i'} \exp(\mathbf{v}_{i'} \mathbf{t}^\top)}
+\tfrac12\,
\frac{\sum_i \mathbf{v}_i \exp(\mathbf{a}_i \mathbf{t}^\top)}
     {\sum_{i'} \exp(\mathbf{a}_{i'} \mathbf{t}^\top)} \\
&= \tfrac12\,\text{Attn}(\mathbf{t},\mathbf{V},\mathbf{V})
   + \tfrac12\,\text{Attn}(\mathbf{t},\mathbf{A},\mathbf{V}).
\end{align}
Similarly, we obtain
\begin{align}
\mathbb{E}[\mathbf{A}\mid\mathbf{t}]
\approx \tfrac12\,\text{Attn}(\mathbf{t}, \mathbf{V}, \mathbf{A})
+ \tfrac12\,\text{Attn}(\mathbf{t}, \mathbf{A}, \mathbf{A}).
\end{align}

\section{Algorithm}
A detailed description of Q-TriM is given in Algorithm~\ref{alg}.

{\SetAlgoNoLine
\begin{algorithm}[t]
\caption{Q-TriM}
\label{alg}
\DontPrintSemicolon

\KwIn{Frame-level video embeddings $\mathbf{V}=\{\mathbf{v}_t\}_{t=1}^T$,
patch-level video tokens $\mathbf{V}^p=\{\mathbf{v}^{p}_{t}\}_{t=1}^T$,
audio embeddings $\mathbf{A}=\{\mathbf{a}_t\}_{t=1}^T$,
question embedding $\mathbf{t}$, keep ratio $k$, number of patches $P$}
\KwOut{Answer logits $\hat{\mathbf{y}} \in \mathbb{R}^C$}

\BlankLine
\textbf{1. Token Filtering}\;
$K \gets kP$\;
\For{$t=1$ \KwTo $T$}{
  $\mathbf{v}^p_t \gets$ Top-$K$ selection of $\mathbf{v}^p_t$\;
}

\BlankLine
\textbf{2. Self-Attention Refinement}\;
$\mathbf{v}^p_t \leftarrow \text{Attn}(\mathbf{v}^p_t,\mathbf{v}^p_t,\mathbf{v}^p_t)$\;
$\mathbf{v}_t \leftarrow \text{Attn}(\mathbf{v}_t,\mathbf{v}_t,\mathbf{v}_t)$\;
$\mathbf{a}_t \leftarrow \text{Attn}(\mathbf{a}_t,\mathbf{a}_t,\mathbf{a}_t)$\;
$\mathbf{t} \leftarrow \text{Attn}(\mathbf{t},\mathbf{t},\mathbf{t})$\;

\BlankLine
\textbf{3. Tri-Modal Attention and Cross Attention}\;
$\mathbf{r}_1 \leftarrow \text{Attn}(\mathbf{t},\mathbf{V}^p,\mathbf{V}^p)$\;
$\mathbf{r}_2 \leftarrow \text{Attn}(\mathbf{t},\mathbf{A},\mathbf{A})$\;
$\mathbf{r}_3 \leftarrow \text{Attn}(\mathbf{t},\mathbf{A},\mathbf{V}^p)$\;
$\mathbf{r}_4 \leftarrow \text{Attn}(\mathbf{t},\mathbf{V},\mathbf{A})$\;

\BlankLine
\textbf{4. Gating}\;
$\mathbf{q} \leftarrow W_g \mathbf{t} + \mathbf{b}_g$,\quad
$g_i \leftarrow 1 + \tfrac12 \tanh(q_i)$\;
$\mathbf{h}_i \leftarrow g_i\,\mathbf{r}_i,\quad i=1,\dots,4$\;

\BlankLine
\textbf{5. Reasoning}\;
$\mathbf{H} \gets [\,\mathbf{t};\,\mathbf{h}_1;\,\mathbf{h}_2;\,\mathbf{h}_3;\,\mathbf{h}_4\,] \in \mathbb{R}^{5\times D}$\;
$\mathbf{F} \gets \text{Attn}(\mathbf{H},\mathbf{H},\mathbf{H})$\;
$\hat{\mathbf{y}} \gets \text{classification layer}(\mathbf{F}) \in \mathbb{R}^C$\;

\Return $\hat{\mathbf{y}}$\;
\end{algorithm}
}

\section{Additional Ablation Study}
We provide an additional ablation study on the MUSIC-AVQA dataset~\cite{MUSIC-AVQA}. Table~\ref{table7} shows the results. We observe a trend similar to that on MUSIC-AVQA-R~\cite{MUSIC-AVQA-R}: each component of Q-TriM contributes to the overall performance.

\begin{table}[h!]
\setlength{\tabcolsep}{2.0pt}
\centering
\caption{Ablation study of Q-TriM on MUSIC-AVQA dataset~\cite{MUSIC-AVQA}.}
\label{table7}

{\small
\renewcommand{\arraystretch}{1.2}
\begin{tabular}{c|cccc}
\Xhline{0.8pt}
\makecell{Method} & A-QA & V-QA & AV-QA & Avg \\
\hline
\textit{w/o} All Modules (Base) & 73.56 & 84.23 & 70.86 & 74.88 \\
\textit{w/o} Token Filtering & 76.60 & 82.58 & 71.94 & 75.58 \\
\textit{w/o} Tri-Modal Attention & 76.78 & 82.78 & 71.47 & 75.41 \\
\makecell[c]{\textit{w/o} Tri-Modal Attention\\(replace with cross attention)}
& 78.40 & 84.72 & 72.00 & 76.50 \\
\textit{w/o} Modality Mixture & 74.55 & 82.62 & 71.55 & 75.01 \\
Ours & \textbf{77.84} & \textbf{85.76} & \textbf{73.78} & \textbf{77.68} \\
\Xhline{0.8pt}
\end{tabular}
}
\end{table}

\section{Qualitative Analysis of Token Filtering}
\begin{figure}[!p]
  \centering
  \begin{subfigure}[b]{\linewidth}
    \centering
    \includegraphics[width=0.9\linewidth]{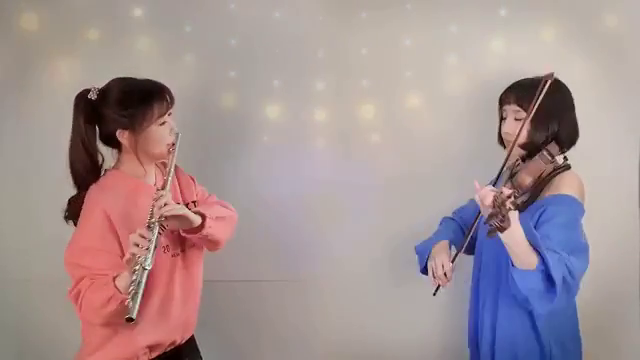}
    \caption{Before filtering}
  \end{subfigure}

  \vspace{0.4em}

  \begin{subfigure}[b]{\linewidth}
    \centering
    \includegraphics[width=0.9\linewidth]{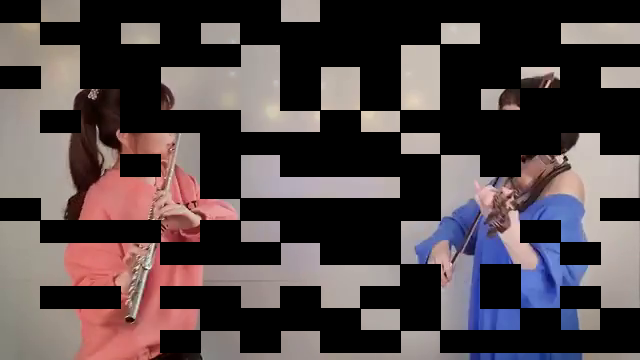}
    \caption{After filtering}
  \end{subfigure}

  \captionsetup{
    format=plain,
    labelsep=period,
    justification=justified,
    singlelinecheck=false,
    width=\linewidth
  }

  \caption{Visualization of token filtering for Example~1. For the question \texttt{"What kind of musical instrument is it?"}, the regions corresponding to the flute and violin are mostly preserved, whereas potentially irrelevant regions are masked out.}
  \label{mask1}
\end{figure}

\begin{figure}[!p]
  \centering
  \begin{subfigure}[b]{\linewidth}
    \centering
    \includegraphics[width=0.9\linewidth]{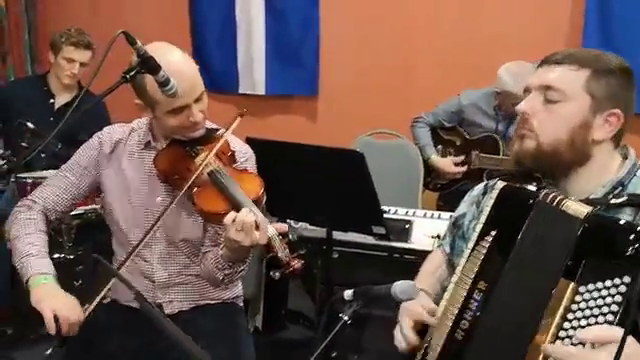}
    \caption{Before filtering}
  \end{subfigure}

  \vspace{0.4em}

  \begin{subfigure}[b]{\linewidth}
    \centering
    \includegraphics[width=0.9\linewidth]{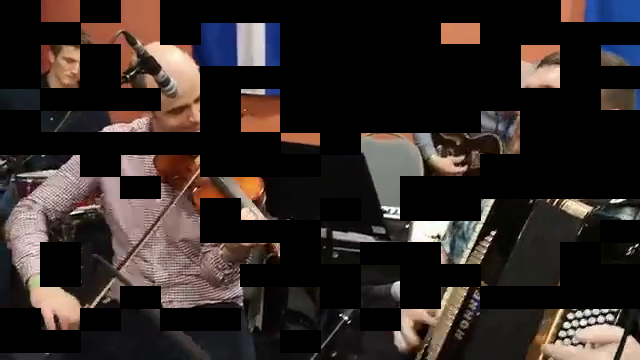}
    \caption{After filtering}
  \end{subfigure}

  \captionsetup{
    format=plain,
    labelsep=period,
    justification=justified,
    singlelinecheck=false,
    width=\linewidth
  }

  \caption{Visualization of token filtering for Example~2.}
  \label{mask2}
\end{figure}

\begin{figure}[!p]
  \centering
  \begin{subfigure}[b]{\linewidth}
    \centering
    \includegraphics[width=0.9\linewidth]{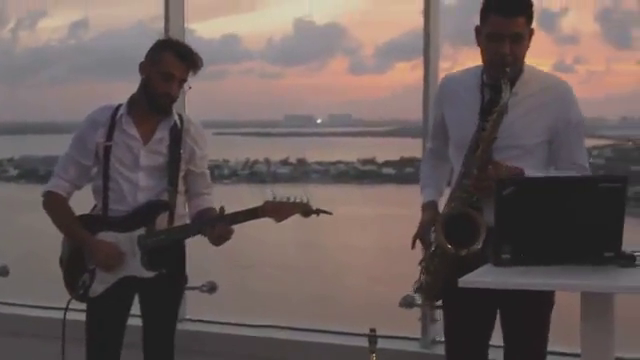}
    \caption{Before filtering}
  \end{subfigure}

  \vspace{0.4em}

  \begin{subfigure}[b]{\linewidth}
    \centering
    \includegraphics[width=0.9\linewidth]{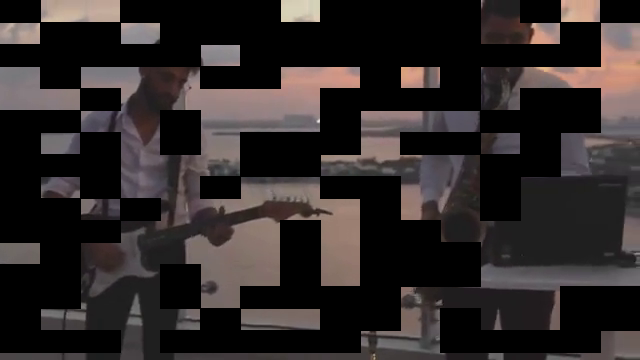}
    \caption{After filtering}
  \end{subfigure}

  \captionsetup{
    format=plain,
    labelsep=period,
    justification=justified,
    singlelinecheck=false,
    width=\linewidth
  }

  \caption{Visualization of token filtering for Example~3.}
  \label{mask3}
\end{figure}

To verify the effectiveness of the token filtering module, we visualize input frames before and after token filtering. The question is:
\[
\texttt{"What kind of musical instrument is it?"}
\]
For three different examples, we show the masked outputs. As shown in Figures~\ref{mask1}--\ref{mask3}, the regions corresponding to musical instruments are more likely to remain after filtering, while potentially irrelevant regions are masked out. This indicates that the token filtering module preserves informative tokens relevant to the question.

\newpage

\section{Adaptive keep-ratio in token filtering}
\color{black}
In general, it is difficult to make $K$ directly learnable/adaptive in \textbf{top-$K$} selection operator for token filtering, because the \emph{hard} top-$K$ operation is non-differentiable preventing the gradient flow to $K$.
% \color{blue}
% More specifically, making $K$ adaptive is difficult because $K$ is an integer-valued variable that controls the \emph{cardinality} of the selected token set. In hard top-$K$ selection, changing $K$ does not produce a smooth change in the output; instead, it changes the size of the selected set discretely and may abruptly alter the selected indices themselves. Therefore, the operator is non-differentiable with respect to $K$, and standard backpropagation does not provide a direct gradient signal for optimizing it. 
% Although straight-through estimation (STE) can be used to route surrogate gradients to the token \emph{scores} under a fixed top-$K$ budget (as in our score-based token selection), this only helps optimize \emph{which} tokens are selected for a given $K$. It does not make the discrete cardinality parameter $K$ itself differentiable, nor does it directly solve the problem of learning \emph{how many} tokens should be retained. 
% \color{black}
Instead, we adopt the workaround proposed in \cite{KEEPRATIO} that uses \emph{soft} top-$K$ selection. Specifically, we \textbf{adjust the temperature parameter} in the softmax approximating hard max-$K$ as follows.
\color{black}
%Rather than changing the cardinality of the selected set itself, we keep the hard top-$K$ operator in the forward pass and make the \emph{soft weighting} around that operator adaptive in the backward pass. In this way, the model preserves the computational efficiency and deterministic sparsification behavior of hard top-$K$, while still allowing gradients to reflect how strongly each token should influence learning.

Let $\mathbf{s}_t \in \mathbb{R}^{P}$ denote the token scores at frame $t$, where $P$ is the number of patch tokens. We define the soft assignment over tokens as
\[
\mathbf{w}_t
=
\mathrm{softmax}\!\left(\frac{\mathbf{s}_t}{\tau}\right).
\]
Here, the temperature parameter $\tau$ controls the sharpness of the distribution. When $\tau$ is small, the softmax becomes sharper and places most of its mass on a few high-score tokens; when $\tau$ is large, the distribution becomes smoother and gradients are spread across a broader set of tokens. Therefore, decreasing $\tau$ has an effect similar to reducing the effective keep ratio, whereas increasing $\tau$ has an effect similar to retaining more tokens during training.

To make the selection input-dependent, we predict $\tau$ from the question and the visual content. Let $\mathbf{t}\in\mathbb{R}^{D}$ be the question embedding and  $\bar{\mathbf{v}}\in\mathbb{R}^{D}$ be a pooled visual representation, e.g., the mean of frame-level video embeddings. We concatenate them and feed the result into a lightweight MLP:
\[
z = [\,\mathbf{t};\bar{\mathbf{v}}\,], \qquad
\tilde{\tau} = \mathrm{MLP}(z).
\]
\noindent\textbf{Brief explanation on MLP} \color{black}
Since the temperature should remain in a stable positive range, we map it into a bounded interval:
\[
\tau
=
\tau_{\min}
+
(\tau_{\max}-\tau_{\min})\,\sigma(\tilde{\tau}),
\]
where $\sigma(\cdot)$ is the sigmoid function, and $\tau_{\min},\tau_{\max}>0$ are predefined bounds. This parameterization allows the model to adapt the softness of token selection according to the input question and scene content, while preventing numerically unstable temperature values.

Under this design, %the hard top-$K$ indices are still obtained from the original scores, but 
the gradient signal is modulated through the temperature-controlled soft weights. In effect, the model does not learn a discrete $K$ directly; instead, it learns how \emph{concentrated} or \emph{diffuse} the token importance distribution should be for each sample. This serves as a continuous surrogate for adaptive keep-ratio control.
\color{black}
Specifically, $\tau$ controls the strength of gradient flow to tokens according to their scores \cite{KEEPRATIO}. A smaller $\tau$ concentrates gradients on high-score tokens, effectively giving a smaller keep ratio.
\color{black}
Conversely, a larger $\tau$ distributes gradients more evenly over tokens, which corresponds to a softer selection behavior and a larger effective keep ratio. Hence, $\tau$ can be interpreted as a differentiable proxy that governs the selectivity of the token-filtering module. % without altering the discrete top-$K$ structure itself.
\color{black}
%We made $\tau$ learnable by setting it as the output of MLP with text and image input.
\color{black}
This strategy was intended to let the model automatically decide whether a given question requires aggressive filtering or broader evidence retention. For example, questions targeting a single salient object may benefit from a sharper distribution with smaller $\tau$, whereas questions requiring broader contextual cues may prefer a smoother distribution with larger $\tau$.

\color{black}
As shown in Table~\ref{table8}, the adaptive variant performed reasonably well, but slightly underperformed the original.
\color{black}
%Although the adaptive variant remained competitive, its performance was consistently slightly lower than that of the fixed-$k$ baseline. 
One plausible explanation is that accurately predicting an appropriate temperature already requires a strong understanding of the question semantics and multimodal evidence. The model must first understand which visual and auditory cues are important in order to decide how selectively it should retain tokens. However, acquiring such multimodal understanding is itself the main objective of AVQA training, leading to a circular dependency.
% \color{black}
% This is perhaps because the model requires strong multimodal understanding to adaptively decide what to retain. However, such understanding is  the goal of AVQA training, leading to a circular argument.
% \color{blue}
% This creates a circular dependency: adaptive filtering is expected to help multimodal reasoning, but determining the correct degree of filtering already depends on multimodal reasoning ability. 
Another possible explanation is that, in early training stages, when representations are still immature, the predicted temperatures may be noisy or suboptimal, which can weaken the effectiveness of token selection and reduce overall robustness. %It is nonetheless interesting to observe that a simple adaptive method performs competitively with the optimal fixed-ratio method. 
\color{black}

\begin{table}[t]
\centering
\caption{Fixed vs. adaptive keep ratio on MUSIC-AVQA dataset~\cite{MUSIC-AVQA}, MUSIC-AVQA-R dataset~\cite{MUSIC-AVQA-R}, MUSIC-AVQA-2.0 dataset~\cite{MUSIC-AVQA-2.0}}
\label{table8}

\begin{subtable}[t]{0.32\linewidth}
\centering
\captionsetup{justification=centering,singlelinecheck=false}
\small
\setlength{\tabcolsep}{2pt}
\renewcommand{\arraystretch}{1}
\begin{tabular}{c|c}
\Xhline{0.8pt}
Method & Accy \\
\hline
Fixed & \textbf{70.89} \\
Adaptive & 70.57 \\
\Xhline{0.8pt}
\end{tabular}
\caption{MUSIC-AVQA-R}
\end{subtable}
\hfill
\begin{subtable}[t]{0.32\linewidth}
\centering
\captionsetup{justification=centering,singlelinecheck=false}
\small
\setlength{\tabcolsep}{2pt}
\renewcommand{\arraystretch}{1}
\begin{tabular}{c|c}
\Xhline{0.8pt}
Method & Accy \\
\hline
Fixed & \textbf{77.68} \\
Adaptive & 76.02 \\
\Xhline{0.8pt}
\end{tabular}
\caption{MUSIC-AVQA}
\end{subtable}
\hfill
\begin{subtable}[t]{0.32\linewidth}
\captionsetup{justification=centering,singlelinecheck=false}
\centering
\small
\setlength{\tabcolsep}{2pt}
\renewcommand{\arraystretch}{1}
\begin{tabular}{c|c}
\Xhline{0.8pt}
Method & Accy \\
\hline
Fixed & \textbf{78.40} \\
Adaptive & 77.59 \\
\Xhline{0.8pt}
\end{tabular}
\caption{MUSIC-AVQA-2.0\\(train: bias, test: bias)}
\end{subtable}

\vspace{0.6em}

\begin{subtable}[t]{0.32\linewidth}
\centering
\captionsetup{justification=centering,singlelinecheck=false}
\small
\setlength{\tabcolsep}{2pt}
\renewcommand{\arraystretch}{1}
\begin{tabular}{c|c}
\Xhline{0.8pt}
Method & Accy \\
\hline
Fixed & \textbf{76.92} \\
Adaptive & 75.71 \\
\Xhline{0.8pt}
\end{tabular}
\caption{MUSIC-AVQA-2.0\\(train: balance, test: bias)}
\label{tab:8d}
\end{subtable}
\hfill
\begin{subtable}[t]{0.32\linewidth}
\centering
\captionsetup{justification=centering,singlelinecheck=false}
\small
\setlength{\tabcolsep}{2pt}
\renewcommand{\arraystretch}{1}
\begin{tabular}{c|c}
\Xhline{0.8pt}
Method & Accy \\
\hline
Fixed & \textbf{75.31} \\
Adaptive & 74.59 \\
\Xhline{0.8pt}
\end{tabular}
\caption{MUSIC-AVQA-2.0\\(train: bias, test: balance)}
\end{subtable}
\hfill
\begin{subtable}[t]{0.32\linewidth}
\centering
\captionsetup{justification=centering,singlelinecheck=false}
\small
\setlength{\tabcolsep}{2pt}
\renewcommand{\arraystretch}{1}
\begin{tabular}{c|c}
\Xhline{0.8pt}
Method & Accy \\
\hline
Fixed & \textbf{76.29} \\
Adaptive & 75.15 \\
\Xhline{0.8pt}
\end{tabular}
\caption{MUSIC-AVQA-2.0\\(train: balance, test: balance)}
\end{subtable}

\end{table}

\section{Additional Baselines}
We provide comparisons with additional closed- and open-source models on the MUSIC-AVQA dataset~\cite{MUSIC-AVQA}. Table~\ref{table11} shows the results where GPT-4o~\cite{gpt} and Qwen2.5-VL~\cite{qwen} are zero-shot results. Our model performs the best, where the second-best MAVEN~\cite{maven} has a 35× larger model size.

\begin{table}
\centering
\caption{Additional Baselines}
\label{table11}

\setlength{\tabcolsep}{6pt}
\renewcommand{\arraystretch}{1.05}

\begin{tabular}{|c|c|}
\hline
\textbf{Method} & \textbf{Accy} \\
\hline
GPT-4o~\cite{gpt} & 56.06 \\
\hline
Qwen2.5-VL~\cite{qwen} & 57.07 \\
\hline
VideoLLaMA2~\cite{videollama} & 66.85 \\
\hline
MAVEN~\cite{maven} & 76.21 \\
\hline
\textbf{Ours} & \textbf{77.68} \\
\hline
\end{tabular}

\end{table}

\end{document}